\documentclass{article}
\usepackage{spconf,amsmath,graphicx}
\usepackage{booktabs}
\usepackage{amssymb}
\usepackage{url}
\usepackage{balance}

\title{Multi-Reference Neural TTS Stylization with \\Adversarial Cycle Consistency}

\name{Matt Whitehill$^{\star}$ \qquad Shuang Ma$^{\dagger}$ \qquad Daniel McDuff$^{\ddagger}$ \qquad Yale Song$^{\ddagger}$}
\address{$^{\star}$University of Washington \qquad $^{\dagger}$University of Buffalo    \qquad $^{\ddagger}$Microsoft Research}

\usepackage{color}

\begin{document}
%
\maketitle
\begin{abstract}
Current multi-reference style transfer models for Text-to-Speech (TTS) perform sub-optimally on disjoints datasets, where one dataset contains only a single style class for one of the style dimensions. These models generally fail to produce style transfer for the dimension that is underrepresented in the dataset. In this paper, we propose an adversarial cycle consistency training scheme with paired and unpaired triplets to ensure the use of information from all style dimensions. During training, we incorporate \textit{unpaired} triplets with randomly selected reference audio samples and encourage the synthesized speech to preserve the appropriate styles using adversarial cycle consistency. 
We use this method to transfer emotion from a dataset containing four emotions to a dataset with only a single emotion. This results in a 78\% improvement in style transfer (based on emotion classification) with minimal reduction in fidelity and naturalness. In subjective evaluations our method was consistently rated as closer to the reference style than the baseline. Synthesized speech samples are available at: 
\url{https://sites.google.com/view/adv-cycle-consistent-tts}
\end{abstract}
%

\section{Introduction}
\label{sec:intro}

The goal of Text-To-Speech (TTS) synthesis is to generate human-like speech based on a text input. Recently, end-to-end trainable neural networks have become increasingly popular for this task. For example, Tacotron \cite{taco} and Tacotron-2 \cite{taco2} use an encoder-decoder architecture that is trained with pairs of text and audio samples $\langle x_{txt}, x_{aud}\rangle$ and a learning objective that the synthesized speech should faithfully reconstruct $x_{aud}$. 
With the success of neural TTS systems, the current focus has been on \textit{TTS stylization}~\cite{shuangma2019neural}, where the goal is to control the style of speech during the synthesis process. The stylization occurs when the system can generate speech for a given text input in a style that is different from what exists in the training data. An ability to control speech style is crucial for developing natural, human-like TTS systems.

\begin{table}[h!]
\centering

 \begin{tabular}{c c c c c} 
  & \multicolumn{4}{c}{\textbf{Emotion}} \\
 \textbf{Speaker ID} & Neutral & Sad & Angry & Happy \\ [0.5ex] 
 \hline\hline
 Speaker 1 & \checkmark &  &  \\ 
 Speaker 2 & \checkmark & \checkmark & \checkmark & \checkmark \\ 
 \bottomrule
 \end{tabular}
 \caption{We use multiple reference audio clips to control different dimensions of speech style (e.g., speaker ID and emotion). We focus on the scenario of \textit{disjoint} datasets, e.g., only one dataset (Speaker 2) contains samples of different emotions.}
 \label{tab:dd}
\end{table}

We use {\it style dimension} to refer to the category of the given style, such as speaker identity, emotion, or accent, and {\it style class} to refer to a specific type such as speaker1, happy, or Scottish. An audio sample $x_{aud}$ has style class labels for either all the defined style dimensions, e.g., it is from speaker1 with happy emotion and a Scottish accent, or only for a subset of the style dimensions, e.g., it is missing the emotion and accent labels.

Multiple systems exist to model the style of speech~\cite{shuangma2019neural, gst, e2e_emotional_speech}, where a reference audio sample with the desired style is used as a conditioning variable during the TTS process. However, most existing approaches require a large number of text-audio training samples of different style dimensions/classes. They also often fail to generalize to new domains unseen during training. For example, to create speech in different speaker identities and emotion classes using a single model, a dataset containing audio samples for each emotion class and speaker identity is needed, and yet the model could still fail to transfer the emotion style to an unseen speaker. Collection of such datasets is challenging and this limits a timely deployment of large-scale TTS stylization systems.

In this paper, we focus on multi-reference neural TTS stylization with \textit{disjoint datasets}. Disjoint datasets occur when one dataset contains samples of only a single style class for one of the style dimensions. Table~\ref{tab:dd} shows a particular scenario we consider in this paper: we use an internal dataset of North American English with two speakers. The dataset for Speaker 1 contains examples for only a single emotion (Neutral) whereas the dataset for Speaker 2 contains examples of all four emotion classes (Neutral, Sad, Angry, Happy). This represents a minimalistic scenario of the aforementioned issue: a model must be able to learn disentangled representations of the two style dimensions, and properly transfer the knowledge about one dimension (emotion) across another dimension (speaker identity) where no variation of style classes is available. This poses a significant challenge to TTS stylization similar to domain adaptation~\cite{daume2006domain}, yet in a unique scenario of style transfer in the speech signal processing domain.

Previous work on TTS stylization has primarily focused on the transfer of a single style reference audio sample \cite{shuangma2019neural, gst, e2e_emotional_speech,  jia2018transfer}. Those methods are inadequate for disjoint datasets because of their lack of domain adaptation capability. In an extreme case, those methods could, for example, learn to identify the emotion using features from the speaker identity dimension. They could also simply ignore the other style dimension (emotion) entirely and always map Speaker 1 samples to the only available style class (Neutral). 

Recently, Bian et al.~\cite{multi-ref} tackled multi-reference TTS stylization, based on GST-Tacotron~\cite{gst} and an intercross training scheme. They showed successful style transfer on a speaker-prosody multi-reference scenario using a 30-hour corpus with 27 speakers and 5 prosodies. However, their intercross training scheme does not guarantee each combination of style classes is seen during training, causing a missed opportunity to learn disentangled representations of styles and sub-optimal results on disjoint datasets. 

In this paper, we address the challenges of multi-reference style transfer on disjoint datasets by using an adversarial cycle consistency training scheme. Unlike intercross training, our training scheme sweeps across all combinations of style classes via paired and unpaired triplets. This provides disentanglement of multiple style dimensions and classes, enabling our model to transfer style in a more faithful manner than existing methods. Testing on our 40-hour disjoint dataset of 2 speakers and 4 emotions, we observe improved emotional expressiveness in synthesized speech, achieving 98.34\% classification accuracy of emotion, a 78.48\% improvement over the baseline model.

\section{Our Method}

\begin{figure}
  \centering
  \includegraphics[width=.99\linewidth]{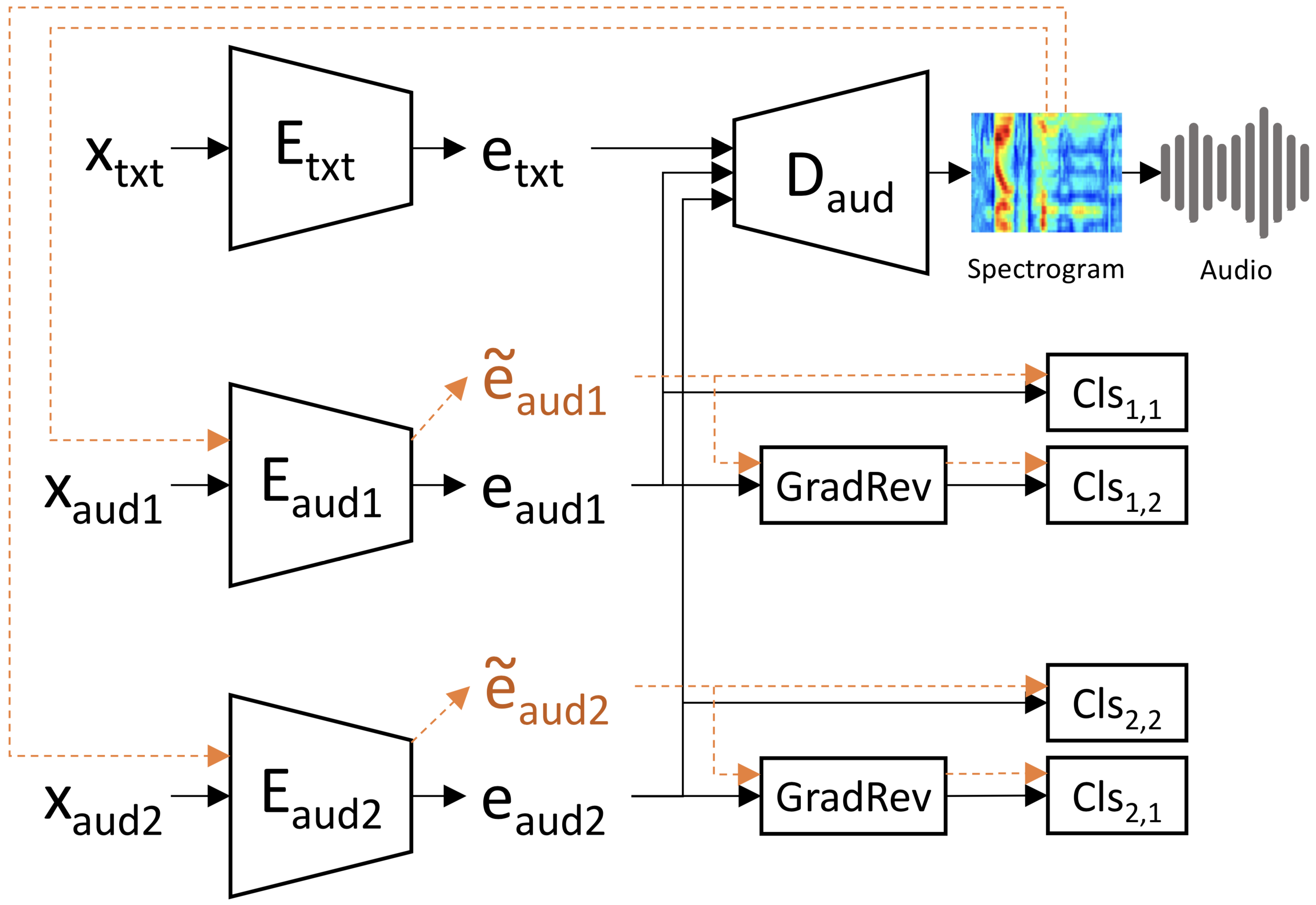}
\caption{Our adversarial cycle consistency training scheme for unpaired samples in a two-reference model. Paired samples are trained with the same scheme and same components, except the synthesized samples $\tilde{e}_{aud}$ are not re-encoded, i.e., the orange dashed lines do not exist for paired samples.}

\label{fig:modelarch}
\end{figure}

Fig~\ref{fig:modelarch} shows a schematic diagram of our system. It consists of a text encoder $E_{txt}$, reference audio encoders $E_{aud1}$ and $E_{aud2}$, and an audio decoder $D_{aud}$. Each audio encoder captures a different style dimension, e.g., $E_{aud1}$ captures speaker identity and $E_{aud2}$ captures emotion. 

At inference time, our model encodes a text string and two reference audio inputs, and produces a spectrogram using the audio decoder; this is converted to the wave file format using the Griffin-Lim vocoder~\cite{griffin1984signal}. More specifically, our text encoder $E_{txt}$ and audio decoder $D_{aud}$ follow the same encoder-decoder architecture of Tacotron-2~\cite{taco2}. We augment this with a reference encoder for each style dimension and concatenate each output embedding $e_{aud1}, e_{aud2}$ to the text context vector at each decoder step. The reference audio encoders follow the same structure as the audio encoder in \cite{tacorefenc}.

During training, we attach style classifiers $Cls_{(\cdot,\cdot)}$ with gradient reversal layers $GradRev$ and feed the generated spectrograms back to the reference audio encoders. This forms the adversarial cycle consistency objective~\cite{zhu2017unpaired}. Below we provide details of our training method. 

\subsection{Model Training}

Learning from disjoint datasets is difficult because we do not have text-audio pairs for all possible combinations of style classes across each style dimension. To encourage disentangling of the style embeddings, we require the model to use both style embeddings $e_{aud1}$ and $e_{aud2}$ during training, with each capturing a different style dimension. Further, we carefully select reference audio samples to ensure each style and speaker is seen during training, filling in the gaps in Table \ref{tab:dd}. 

We achieve this by synthesizing speech from both paired and unpaired triplets. We use the convention $x_{aud_{n,p}}$ to represent reference audio samples, where $n$ stands for the style dimension and $p$ stands for the pairing type. The pairing type can take one of three values: \textbf{1)} a \textit{paired} audio sample $x_{aud_{n,+}}$ with the same verbal content as the input text, \textbf{2)} a \textit{style-matched} audio sample $x_{aud_{n,*}}$ with the same style class as the paired audio sample but with a different verbal content than the input text, \textbf{3)} a \textit{random} audio sample $x_{aud_{n,-}}$ with a random style class. 

A \textit{paired triplet} contains a text sample, a paired audio sample, and a style-matched audio sample, and it can be either $\langle x_{txt}, x_{aud_{1,+}}, x_{aud_{2,*}}\rangle$ or $\langle x_{txt}, x_{aud_{1,*}}, x_{aud_{2,+}}\rangle$. An \textit{unpaired triplet} contains a text sample and two random audio samples, $\langle x_{txt}, x_{aud_{1,-}}, x_{aud_{2,-}}\rangle$. Our style-matched sample is similar to that in the intercross training scheme used by Bian et al.~\cite{multi-ref}, and our random sample is similar to the unpaired training scheme used by Ma et al.~\cite{shuangma2019neural}. In this work, we combine those ideas to enable multi-reference TTS stylization from disjoint datasets. Next, we discuss the loss functions used for the paired and unpaired triplets.

\textbf{Reconstruction loss.} For the paired triplets only, we force the synthesized spectrograms to reconstruct the paired audio sample. We follow \cite{taco,shuangma2019neural,gst} and define an $\mathcal{L}_1$ reconstruction loss between the input spectrogram $x_{aud}$ and the output spectrogram $\tilde{x}_{aud}$,
$$\mathcal{L}_{recon} = \| x_{aud} - \tilde{x}_{aud} \|_1$$

\textbf{Adversarial Cycle Consistency Loss.} 
The reconstruction loss alone is insufficient to constrain our model. Inspired by \cite{zhu2017unpaired}, we introduce an adversarial cycle consistency loss to further constrain it. Our main idea is that an embedding $e_{aud}$ from the real audio sample must capture the correct style information. Thus, when we synthesize audio from it and feed the result back to the same audio encoder, the resulting embedding $\tilde{e}_{aud}$ should contain the same style information as $e_{aud}$; hence the \textit{cycle consistency}. Furthermore, each of the two audio embeddings $e_{aud(\cdot)}$ should only contain information about the corresponding style dimension; in other words, $e_{aud1}$ should have no information about style dimension two, and similarly $e_{aud2}$ should not have information about style dimension one; this can be enforced via \textit{adversarial learning}.

We design our adversarial cycle consistency loss by combining the two ideas above. To this end, we define style classifiers $Cls_{i,j}$ where $i$ refers to the style dimension of the input embedding and $j$ refers to the style dimension upon which the classification occurs. The classifier is a two-layer MLP with a softmax classifier and outputs equal to the number of style classes for the $j$-th style dimension. We train it with a cross-entropy loss:
$$ \mathcal{L}_{cls} = -\sum_{i,j} y_{i,j} \log(\tilde{y}_{i,j})$$
where $y_{i,j}$ is the ground-truth style class for the $i$-th embedding in the $j$-th style dimension, and $\tilde{y}_{i,j}$ is the predicted style class. For $i=j$, the classifier encourages an embedding $e_{aud_{i}}$ to contain the correct information of the $i$-th style dimension. For $i \neq j$, the classifier discourages the use of information about the other style dimension. We use the gradient reversal layer~\cite{ganin15unsupervised} before the classifiers for $i \neq j$ to enable adversarial learning.

The adversarial cycle consistency loss is then a combination of classification losses for paired triplets, unpaired triplets, and synthesized samples (with $\delta=.01$), $$ \mathcal{L}_{adv.cycle} = \mathcal{L}_{cls,paired} + \mathcal{L}_{cls,unpaired} + \delta \mathcal{L}_{cls,synthesized}$$

\textbf{Orthogonality loss.} Finally, we introduce an orthogonality constraint to help the model learn disentangled style representations, similar to \cite{multi-ref}. This is defined over the style embeddings as $$ \mathcal{L}_{ortho} = \sum_{i,j} \| e_{aud_{i}}^{\intercal} e_{aud_{j}} \|_{F},$$ where $||\cdot||_F$ is the Frobenius norm and $e_{aud_{i}}$ (and $e_{aud_{j}}$) refers to the style embedding from style dimension $i$ (and $j$).

\textbf{Training details.}
The final form of our loss function is $$\mathcal{L} = \alpha \mathcal{L}_{recon} + \beta \mathcal{L}_{adv.cycle} + \gamma \mathcal{L}_{ortho}$$ where $\alpha= \beta = 1.0$ and $\gamma=0.02$ are weights for the different loss terms. We found the optimal weights and that the results are insensitive to small changes to those values through cross-validation. We train our model on a single machine with four NVIDIA Tesla M40 GPUs for 40k epochs using a batch size of 96 text/audio pairs, each with a paired and unpaired triplet, for a total of 192 triplets. Note that $\mathcal{L}_{recon}$ is defined over only the paired triplets while the other two loss terms are defined over both paired and unpaired triplets. We use teacher-forcing for the reconstruction loss throughout the entire training procedure. At inference time, we use a window constraint for the text context attention, enforcing the maximum attention weight to be within a window of seven frames from the previous max. For the rest of the hyperparameters, we follow the same setup as outlined in \cite{taco2}. 
After the 40k epochs, we add the adversarial game loss presented in Ma et al.~\cite{shuangma2019neural} and train for an additional 1k epochs. Fine-tuning the model with this loss increases the fidelity of the synthesized unpaired samples. 

\section{Experiments and Discussions}
\label{sec:expsetup}
Our disjoint datasets are defined over two style dimensions, speaker identity and emotion, as shown in Table~\ref{tab:dd}. The datasets contain 15,226 samples (18.55 hours) and 22,325 samples (21.62 hours) for Speakers 1 and 2 respectively. To the best of our knowledge, there exists only one published method that tackles multi-reference TTS stylization:
we compare to Bian et al.~\cite{multi-ref} in our experiments.

\textbf{Style Classification Accuracy.}
\label{ssec:styclassacc}
We train two speech style classifiers (speaker identity and emotion) using the reference audio samples from the TTS training data. The classifiers have the same structure as the reference encoder and the style classifier in our model. Their final validation accuracies are 99\% and 95\% respectively.

Next, we synthesize speech from each test text sample four times, once in each emotion, and predict their style class labels using the trained classifiers. For the emotion reference, we use a random sample in the appropriate emotion from the Speaker 2 test set. For the speaker identity reference, we use the paired audio sample. 

\begin{figure}[htb]
\begin{minipage}[b]{.49\linewidth}
  \centering
  \centerline{\includegraphics[width=4.4cm]{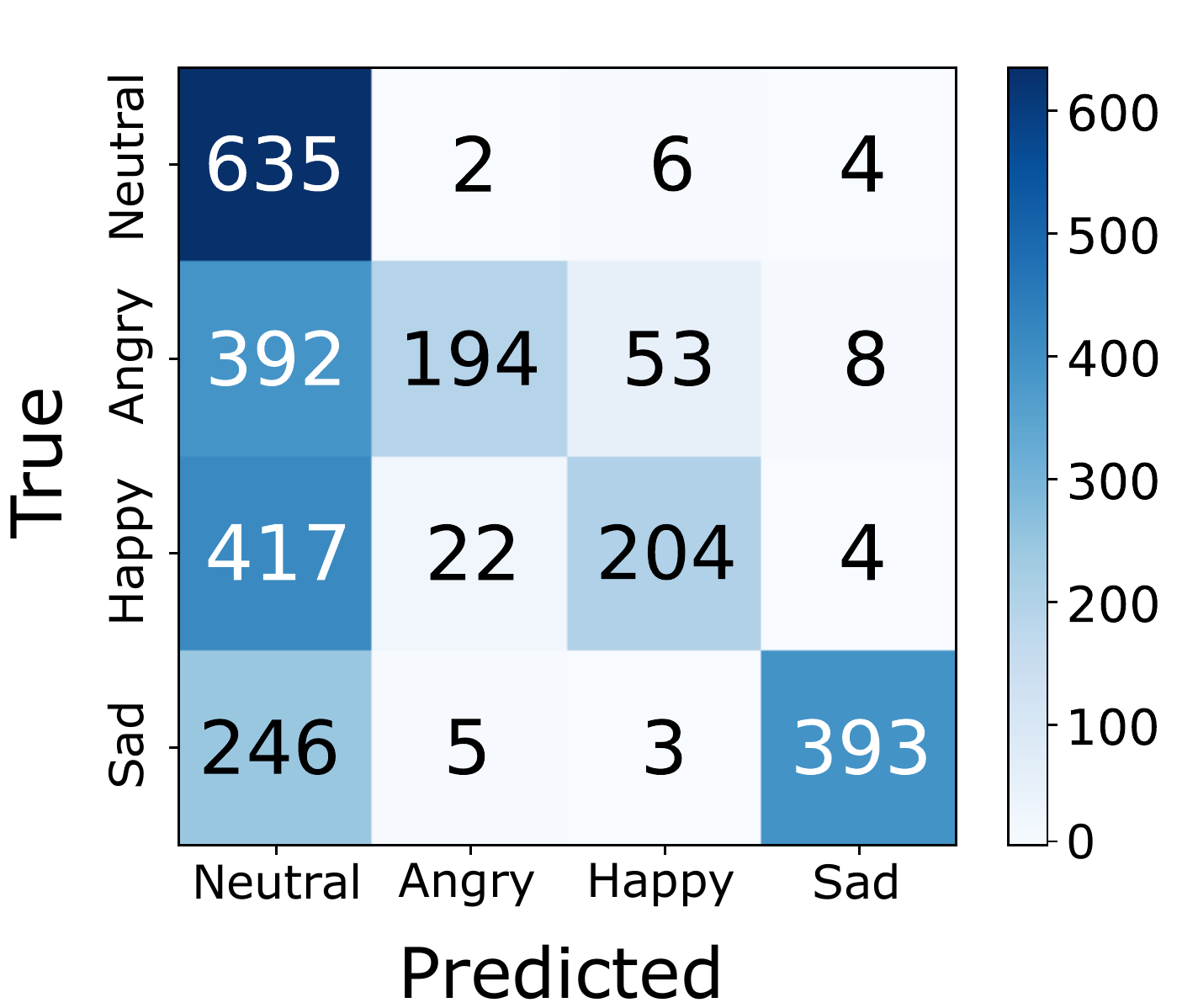}}
  \centerline{(a) Bian \textit{et al.}~\cite{multi-ref}}\medskip
\end{minipage}
\hfill
\begin{minipage}[b]{.49\linewidth}
  \centering
  \centerline{\includegraphics[width=4.4cm]{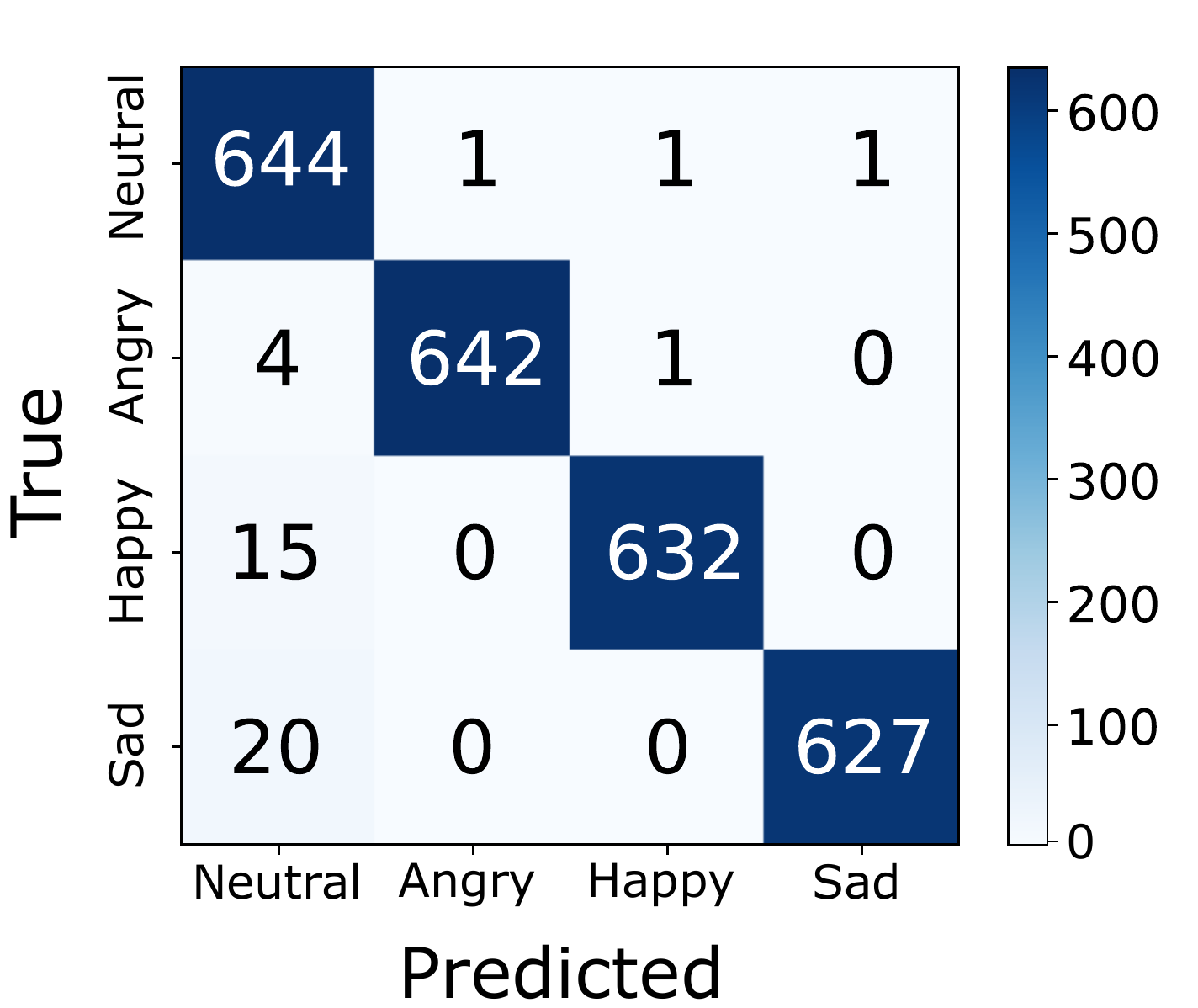}}
  \centerline{(b) Our Model }\medskip
\end{minipage}
\caption{Confusion matrix for emotion classification results of Speaker 1 synthesized samples.}
\label{fig:cnfmatrix}
\end{figure}

Table \ref{tab:classresults} shows the classification results and Figure \ref{fig:cnfmatrix} shows the confusion matrices. Both models achieve greater than 96\% accuracy on speaker identity, showing their ability to retain speaker identity in synthesized samples. However, the baseline model performs poorly on emotion classification, achieving only a 55.1\% classification accuracy. As can be seen in the confusion matrix, many samples from the angry, happy, and sad classes are grouped into the neutral class, demonstrating a lack of style transfer. Our model achieves 98.3\% classification accuracy, demonstrating a much higher rate of emotion style transfer.

\begin{table}[htb]
\centering
 \begin{tabular}{c | c c} 
   & \textbf{Emt Acc (\%)} & \textbf{Spk Acc (\%)}\\
 \hline\hline
 Bian \textit{et al.}~\cite{multi-ref}  &  55.1 & 97.1 \\
 Our Model  & 98.3 & 96.9 \\
 \bottomrule
 \end{tabular}
 \caption{Results of style classification for emotion (Emt) and speaker identity (Spk). We use only Speaker 1 text samples for the emotion classification.}
 \label{tab:classresults}
\end{table}


\begin{figure}[htb]
\begin{minipage}[b]{.49\linewidth}
  \centering
  \centerline{\includegraphics[width=4.4cm]{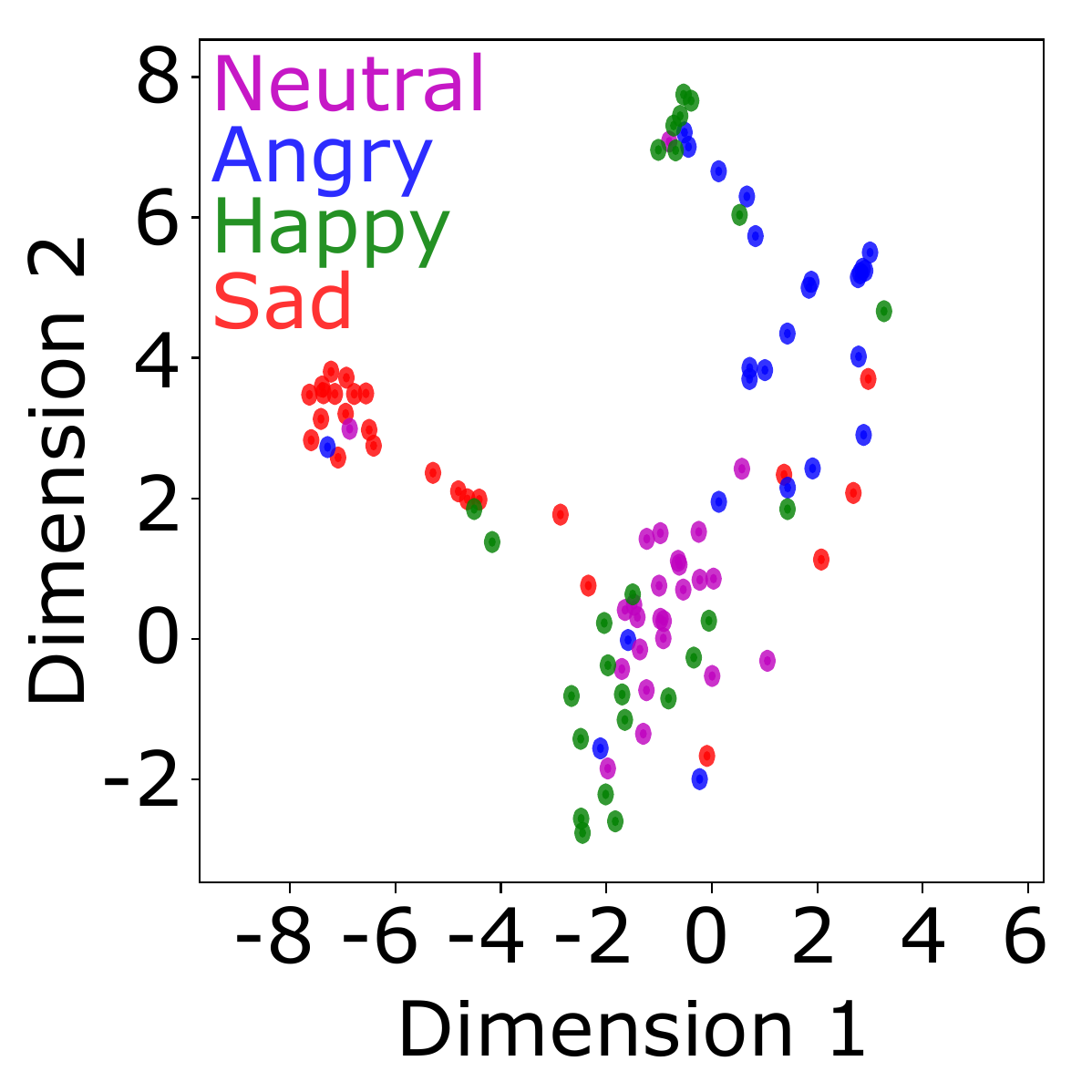}}
  \centerline{(a) Bian \textit{et al.}~\cite{multi-ref}}\medskip
\end{minipage}
\hfill
\begin{minipage}[b]{.49\linewidth}
  \centering
  \centerline{\includegraphics[width=4.4cm]{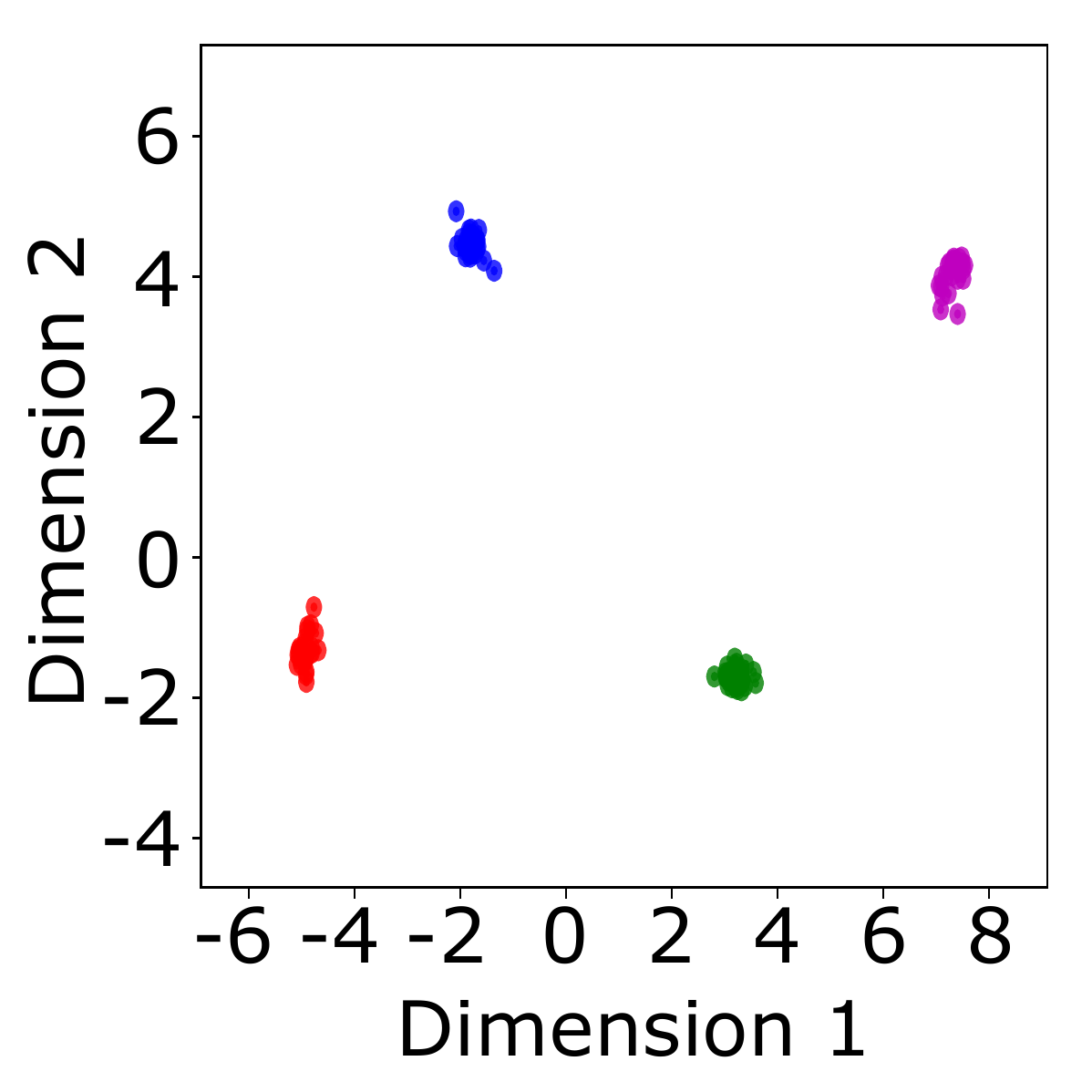}}
  \centerline{(b) Our Model }\medskip
\end{minipage}
\caption{T-SNE plots for the emotion embeddings for our model (b) compared to the baseline (a).}
\label{fig:tsne}
\end{figure}

We also visualize 100 embeddings (25 from each emotion) created by the emotion classifier's reference encoder using t-SNE \cite{maaten2008visualizing} in Figure \ref{fig:tsne}. Our model produces much closer and more separable clusters due to the improved emotion style transfer; the results suggest an improved disentanglement of the two style dimensions using our model.

\textbf{Human Subject Evaluation.} We recruited eight human subjects to qualitatively evaluate our adversarial cycle consistency model. To test style transfer, we performed a side-by-side comparison of 20 synthesized Speaker 1 samples (5 texts in each of the 4 emotions). Subjects evaluated the samples on a 7-point scale (-3 to 3) where -3 refers to ``sample A is closest to the reference emotion''. The results show our model was consistently rated as closer to the reference ($\mu=0.86$), especially for the three unseen emotions in the Speaker 1 dataset (sad: $\mu=2.03$, angry: $\mu=0.98$, happy: $\mu=0.63$).


To test naturalness, we asked subjects to rate voice quality on a 5-point scale. Our model achieved a 3.29 mean opinion score (MOS) while the baseline reached a 3.43 MOS. Our model's reduction in perceived quality may result from its more pronounced style transfer -- on neutral samples, our model (3.63 MOS) outperforms the baseline (3.40 MOS). Perhaps the style transfer was too strong (almost exaggerated) for the other three emotions, leading to a decrease in the naturalness score.

\textbf{Speech Fidelity.}
\label{ssec:speechfidelty}
Finally, we evaluated the fidelity of the synthesized speech samples. We synthesized each Speaker 1 test text sample in each of the four emotions, then use the Microsoft Azure speech-to-text service to generate transcripts. The baseline reaches 15.75\% word error rate (WER) while our model achieves 16.95\%. Similar to naturalness, our model's improved emotional expressiveness may be the cause of its lower performance since the emotion can serve to confound the automatic speech recognition system. We also believe that improved fidelity could be achieved with a more powerful vocoder such as WaveNet \cite{oord2016wavenet}.  

\textbf{Comparison with  Bian et al.~\cite{multi-ref}}. We believe the baseline model's sub-optimal performance stems from the limitations of the intercross training procedure. Since the procedure only presents combinations of style classes that exist in the dataset (e.g. entries with a check-mark in Table \ref{tab:dd}), unrepresented combinations (e.g. the gaps in Table \ref{tab:dd}) do not impact the model loss and, thus, are not accounted for during backpropagation. By training on unpaired triplets with random references, our cycle consistency training scheme ensures each combination of style class (e.g. each entry in Table \ref{tab:dd}) is seen during training, forcing the model to learn to create speech for every style combination.

\section{Conclusion}
\label{sec:conclusion}
We present an adversarial cycle-consistent training procedure for multi-reference neural TTS stylization on disjoint datasets. Because recording training samples for new style classes is labor-intensive, transferring style from one dataset to another (including disjoint datasets) is an appealing feature for TTS systems. Using our adversarial cycle consistency training scheme, we achieve a much higher rate of style transfer for disjoint datasets than previous models. We show our model provides a 78\% improvement in style transfer (based on emotion classification) over an existing method with minimal reduction in fidelity and naturalness.

\balance{}
\bibliographystyle{IEEEbib}
\bibliography{strings,refs}

\end{document}